\documentclass{article}
\usepackage{ijcai24}

\usepackage{times}
\usepackage{soul}
\usepackage{url}
\usepackage[hidelinks]{hyperref}
\usepackage[utf8]{inputenc}
\usepackage[small]{caption}
\usepackage{graphicx}
\usepackage{amsmath}
\usepackage{amsthm}
\usepackage{amssymb}
\usepackage{booktabs}
\usepackage{color}
\usepackage{multirow}
\usepackage{algorithm}
\usepackage{algorithmic}
\usepackage[switch]{lineno}

\title{SFPrompt: Communication-Efficient Split Federated Fine-Tuning for Large Pre-Trained Models over Resource-Limited Devices}

\author{
Linxiao Cao$^1$
\and
Yifei Zhu$^2$\And
Wei Gong$^{1}$
\affiliations
$^1$University of Science and Technology of China\\
$^2$Shanghai Jiao Tong University\\
\emails
linxiaocao@mail.ustc.edu.cn,
yifei.zhu@sjtu.edu.cn,
weigong@ustc.edu.cn
}

\begin{document}

\maketitle

\begin{abstract}
Large pre-trained models have exhibited remarkable achievements across various domains. 
The substantial training costs associated with these models have led to wide studies of fine-tuning for effectively harnessing their capabilities in solving downstream tasks. Yet, conventional fine-tuning approaches become infeasible when the model lacks access to downstream data due to privacy concerns. Naively integrating fine-tuning approaches with the emerging federated learning frameworks incurs substantial communication overhead and exerts high demand on local computing resources, making it impractical for common resource-limited devices. In this paper, we introduce SFPrompt, an innovative privacy-preserving fine-tuning method tailored for the federated setting where direct uploading of raw data is prohibited and local devices are resource-constrained to run a complete pre-trained model. In essence, SFPrompt judiciously combines split learning with federated learning to handle these challenges. Specifically, the pre-trained model is first partitioned into client and server components, thereby streamlining the client-side model and substantially alleviating computational demands on local resources. SFPrompt then introduces soft prompts into the federated model to enhance the fine-tuning performance. To further reduce communication costs, a novel dataset pruning algorithm and a local-loss update strategy are devised during the fine-tuning process. Extensive experiments demonstrate that SFPrompt delivers competitive performance as the federated full fine-tuning approach while consuming a mere 0.46\% of local computing resources and incurring 53\% less communication cost.
\end{abstract}

\section{Introduction}
Large pre-trained models have achieved unprecedented success across various domains, including natural language processing (NLP) and computer vision (CV) \cite{brown2020language,wang2023large}. 
To support more diverse and complex tasks, the size of the pre-trained models has increased substantially, e.g., the model sizes of the GPT series have increased from 117M to 175B \cite{desislavov2021compute}. 
Nevertheless, as the model size increases, so too does the training cost (e.g., training GPT-3 would cost over \$4.6M \cite{dale2021gpt}). 
To lower training costs and enable wide adoption, a prevailing paradigm is to fine-tune the pre-trained models to adapt downstream tasks.
However, existing fine-tuning methods often necessitate access to downstream task data, which is usually not possible in practice.
With the enactment of regulations like the EU's General Data Protection Regulation (GDPR) \cite{voigt2017eu}, along with increasing attention to privacy and security, obtaining downstream data becomes more and more challenging. How to conduct fine-tuning without accessing raw downstream data is a fundamental problem that needs to be solved. 

To work in these privacy-preserving environments, integrating fine-tuning with the emerging federated learning (FL) \cite{zhao2023fedprompt,xiao2023offsite,chen2022fedtune} has gained traction. However, existing efforts still have the following limitations:

\begin{figure}
    \centering
    \includegraphics[width=0.95\linewidth]{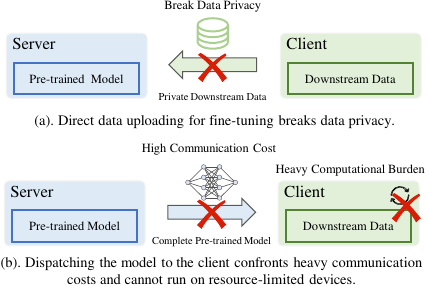}
\caption{Limitations of current fine-tuning approaches in privacy-preserving and resource-limited environments}
    \label{motivation}
\end{figure}

\textbf{Overwhelming Communication Cost.} As the model size increases, collaboratively fine-tuning a pre-trained model introduces significant communication costs due to model exchange. 
For instance, GPT-3 \cite{brown2020language}, encompassing a staggering 175B parameters that total approximately 650GB, renders the task of uploading and downloading virtually infeasible.

\textbf{Substantial Local Computational Load.} Existing federated fine-tuning studies assume that the downstream users also possess the pre-trained model. The resulting local fine-tuning thus becomes intensely computation-demanding, which places stringent requirements on the local device's computation power. Continuing with GPT-3 as an example, its 175B parameters require 350GB GPU memory for parameter storage \cite{xiao2023offsite}, making it challenging to handle on consumer-grade hardware like the NVIDIA RTX 3090 (with 24GB of memory)\footnote{https://www.nvidia.com/en-us/geforce/graphics-cards/30-series/rtx-3090-3090ti}. Besides, based on an inference task involving 1M tokens, GPT-3 demands 740 PFLOPs, translating to around 30 days of computation on an 8-card RTX 3090 setup \cite{desislavov2021compute}.

To address these challenges, we present SFPrompt, a distributed fine-tuning framework that can adapt large pre-trained models to downstream tasks without accessing to the raw scattered data in resource-constrained environments. 
SFPrompt divides the model into three segments: the head for extracting features from the inputs, the body for capturing feature interdependencies, and the tail for mapping the features to task-specific output. 
We streamline the model between clients and the server by allowing the head and the tail to run on the client side, and the body on the server side, which greatly reduces the demand on local computing resources. 
We prune the local dataset by filtering out unimportant samples to send less data in each interaction and update the model based on local loss to reduce the number of interactions. In this process, the raw data is always stored on the local side without sharing, providing privacy preservation. In addition, we introduce learnable prompt parameters to modify the model’s input space, enabling efficient fine-tuning. Our key contributions are outlined as follows:


\begin{itemize}
\item SFPrompt is the first work to fine-tune a large pre-trained model in a privacy-demanding and resource-constrained environment, where raw data are not allowed to be aggregated and local computing devices have low computation power. 
\item We split the large pre-trained model into server-side and client-side to lower the computational burden on the client and further introduce prompt parameters to enable efficient fine-tuning.
\item We prune the redundant local data and update the model on the client side based on the local loss to further reduce the additional communication cost.
\item The comprehensive experiments demonstrate SFPrompt can achieve competitive performance to existing methods while consuming only 0.46\% of local computing resources and incurring 53\% less communication cost.
\end{itemize}

\section{Background and Related Work}

\subsection{Federated Fine-tuning Pre-trained Models}
To adapt large pre-trained models to specific downstream tasks efficiently, researchers have proposed a variety of parameter-efficient fine-tuning methods \cite{houlsby2019parameter,hu2021lora,jia2022visual}. However, these methods all assume complete access to the raw data, which is not always possible with the growing awareness of data privacy and the enforcement of data regulations. 
Consequently, recent works applied the FL framework to fine-tuning pre-trained models in privacy-preserving environments. The existing methods can be broadly categorized into two main streams. The first involves a direct combination of FL and fine-tuning methods \cite{zhao2023fedprompt,guo2023promptfl,zhang2022federated}. For example, Zhao et al. \cite{zhao2023fedprompt} freezes the pre-trained model and aggregates prompts to fine-tune the pre-trained models. The second approach focuses on combining parameter-efficient fine-tuning and model emulator derived from a lossy compressed version of a large pre-trained model. Xiao et al. \cite{xiao2023offsite,niu2022federated} fine-tunes the adapter on the downstream data with the emulator’s assistance. 
Although these early studies confirm the potential of FL in fine-tuning pre-trained models, they still demand high computational local devices and incur significant communication costs. 



\subsection{Split Federated Learning}
Split federated learning (SFL) \cite{thapa2022splitfed} is a novel framework 
that combines Split Learning (SL) \cite{gupta2018distributed,vepakomma2018split} and FL \cite{mcmahan2017communication,konevcny2016federated,kairouz2021advances}. SFL reduces local computing costs by using SL to split the model into two modules: the client model and the server model. The client model has lower computational complexity, minimizing computational overhead at clients. The server model with heavy complex computations is shouldered by the server. SFL then achieves faster collaboratively training by performing parallel processing across clients using FL. Intuitively, SFL can be a good framework to address the training challenges in distributed scenarios with constrained local resources.

\begin{figure}[t]
\centering
\includegraphics[width=\linewidth]{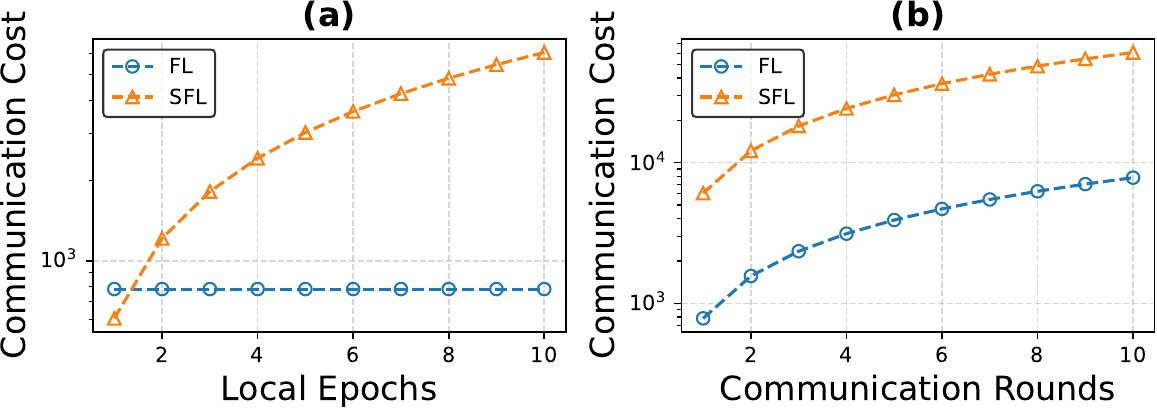}
\caption{(a) communication cost comparison in a global round between FL and SFL; (b) the relationship between the communication rounds and communication cost.}
\label{Rounds VS. Cost}
\end{figure}

\begin{figure*}
    \centering    
    \includegraphics[width=1\linewidth]{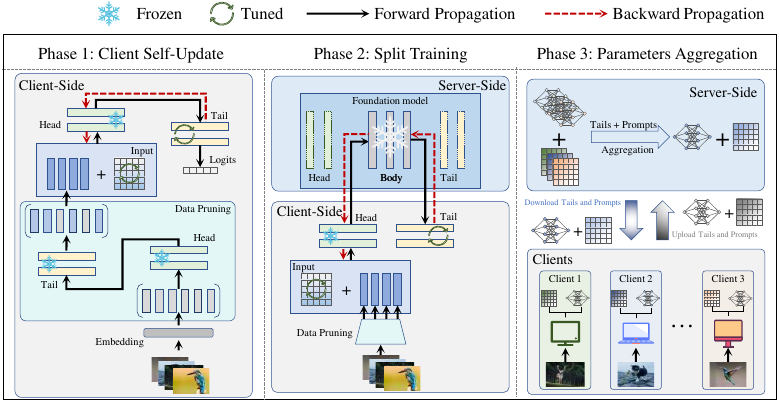}
    \caption{
    The overview of SFPrompt.
    SFPrompt is a three-phase process: Phase 1 minimizes communication cost by implementing local-loss updates and dataset pruning; Phase 2 splits the foundation model to reduce local computational burdens by remaining the lightweight part on client;  Phase 3 aggregates the tail model and prompt parameters to refresh the global model.
    }
    \label{overview}
\end{figure*}

Nevertheless, naively applying SFL to fine-tune large pre-trained models still introduces significant communication burdens because of the frequent transmission of forward and backward signals. 
To further validate the point, we use a pretrained vision transformer \cite{dosovitskiy2020image} as an example.
Suppose there is a client with 1,000 locally available images. During each global round, which represents the cycle from the current parameter aggregation to the next, the client performs 10 local epochs, representing the number of local model updates within a single global round.
Figure \ref{Rounds VS. Cost} details the communication cost of FL and SFL. It's clear that the SFL's communication cost is much higher than traditional FL methods. This discrepancy arises because, within each global round, FL needs to transmit the model twice. Meanwhile, during each local epoch of a global round, SFL is required to transmit the smashed data and gradients. Although the cost is lower than FL in initial epochs, as depicted in Fig \ref{Rounds VS. Cost} (a), the communication cost becomes intolerable as the local epochs increase.
This issue calls for a careful design of SFL for pre-trained model fine-tuning so that the communication and computation costs can be reduced in a practical privacy-preserving and resource-constrained environment.

\section{Methodology}
\subsection{Framework Overview}
We propose SFPrompt for efficiently adapting large pre-trained models to distributed downstream tasks with private data. SFPrompt employs a methodical three-phase strategy: client self-update, split training, and parameters aggregation, executed sequentially to ensure optimal performance. We illustrate the overall framework of SFPrompt in Figure \ref{overview}.


SFPrompt partitions the model, denoted as $W$, into three components to fully leverage the local computational resource. The head is the first few layers represented as $W_{h}$ for feature extracting, the body is represented as $W_{b}$ for modeling the dependency between features, and the tail is the classifier represented as $W_{t}$ for mapping the feature to the task-specific output. In practical terms, the splitting strategy in SFPrompt is dynamic, not rigid. It is determined by the client's hardware resources, ensuring optimal performance without overwhelming the client's hardware.

The server distributes the client-side model, formed by the head and tail models and denoted as $W_{C}=[W_{h}, W_{t}]$, to the selected clients $K$, where $k\in K$ denotes the $k_{th}$ client participating in the training process. Conversely, the body $W_{b}$ is housed on the server, forming $W_{S}$. $W_{C}$ exhibits a lower computational complexity and model size compared to the foundation model $W$, making it more suitable for implementation at clients, particularly in resource-limited scenarios.

\subsection{Phase 1: Client Self-Update}
Upon careful observation in the related work, which is shown in Fig \ref{Rounds VS. Cost}, it becomes apparent that merely employing SFL, coupled with the existing fine-tuning methods, engenders new challenges. To tackle this issue, we approach the problem from two angles in SFPrompt. 

\noindent\underline{\textbf{Local-loss Update.}} To reduce the number of interactions between the server and the client, SFPrompt introduces the local-loss to achieve client self updating. 
We establish a connection between the final layer of $W_{h,k}$ and the local tail $W_{t,k}$, such that the output of $W_{h,k}$ becomes the input of $W_{t,k}$. We define the local loss function as $\mathcal{L}_{C}$:
\begin{equation}
\mathcal{L}_{C} = \frac{1}{|D_{k}|}\sum_{x \in D_{k}} \ell(x;(W_{h,k},W_{t,k});p)
\end{equation}
where $l(x;(W_{h},W_{t});p)$ is the loss computed based on the input $x$, the head model $W_{h,k}$, the local tail model $W_{t,k}$ and the prompt $p$. By inputting private data into the constructed model to perform local-loss updates, we eliminate the need for frequent interaction with the server compared with SFL, thereby reducing communication costs. 

During this process, the local tail model $W_{t}$ and the prompt $p$ updates, while the head model remains frozen. Importantly, this process incurs no additional communication costs as it does not establish a connection to the server-side model.

\noindent\underline{\textbf{Local Dataset Pruning.}} Although we've made efforts to minimize the overall number of interactions, the cost of each individual interaction remains high.
Before the split training, the selected clients participating in the training perform dataset pruning. We initially identify a collection of important training samples from the local dataset. We employ the norm of the error vector (EL2N) \cite{paul2021deep}, which is defined as follows: 
\begin{equation}
    \mathbb{E}||p(W_{t}, x)-y||_{2}
\end{equation}
where $p(W_{t}, x)$ represents the output of the neural network in the form of a probability vector when the input is $x$, and $y$ denotes the one-hot encoded labels. EL2N scores are remarkably effective at identifying significant examples, which enables us to gauge the impact of a training point on the loss of an arbitrary example. By focusing on training samples with greater effect, we can reduce the need for unnecessary data transfer. 

Specifically, we link the head $W_{h}$ and the tail ${W_{t}}$, procuring the EL2N scores through the predicted class probability minus the one-hot label encoding. We prune the dataset using a preset pruning fraction $\gamma$, ensuring that we retain the examples with higher EL2N scores. Therefore, the communication process only passes these processed important samples, not all samples, thereby reducing the communication cost.

Overall, SFPrompt minimizes single-round data transmission by filtering local redundant data and leveraging local-loss updates to reduce frequent interactions, effectively cutting communication costs. The more details of SFPrompt on the client can be referred to Algorithm \ref{SFPrompt(client)}. 

\begin{algorithm}[t]\small
  \renewcommand{\algorithmicrequire}{\textbf{Input:}}
  \renewcommand{\algorithmicensure}{\textbf{Output:}}
  \caption{SFPrompt (Client)}
  \label{SFPrompt(client)}
  \begin{algorithmic}
      \REQUIRE $D_{k}=\{z_{1}, z_{2},\ldots,z_{n}\}$ denotes the dataset positioned on client $k$, where each $z_{i}$ is a pair $(x_{i}\in X,y_{i}\in Y)$. The $k_{th}$ client model, denoted as $W_{C,k}$, encompasses the elements $[W_{h,k},W_{t,k}]$. The number of local epochs is represented as $U$.

      \STATE \textbf{\emph{Local Dataset Pruning}}
      \STATE \quad Connect the local network $W_{h}$ and $W_{t}$
      \STATE \quad Encode the labels $\overline Y =$ \textbf{\emph{One-hot}} $(Y)$
      \STATE \quad Calculate the Error $ =$ \textbf{\emph{Softmax}} $(W_{t},(W_{h},X))-\overline Y$
      \STATE \quad EL2N scores $=L_{2}$ norm (Error)
      \STATE \quad Sort $D_{k}$ in descending order based on scores
      \STATE \quad Construct the $\gamma$-pruning subset: 
      \STATE \quad \quad$\hat{D_{k}}=\{z_{i}| \forall z_{i}\in D_{k}, i>\gamma*n\}$
      \STATE \;

      \STATE \textbf{\emph{Local-loss Update}}
      \FOR{local $u\in U$}
      \STATE Connect the local network $W_{h}$ and $W_{t}$
      \STATE Incorporate prompts $p$ to the embedding of the inputs 
      \STATE Update $W_{t,u+1} = W_{t,u}-\eta \nabla \ell (W_{h,u},W_{t,u},p_{u})$
      \STATE Update $p_{u+1} = p_{u}-\eta \nabla \ell (W_{h,u},W_{t,u},p_{u})$
      \ENDFOR
      \STATE return $W_{h,U}$, $W_{t,U}$, $p_{U}$

      \STATE\;
      \STATE \textbf{\emph{Client Forward Update}}
      \STATE \quad $W_{h,k}$, $W_{t,k}$, $p_{k}$ $\longleftarrow$ \textbf{\emph{Local-loss Update}}
      \STATE \quad $\hat{D_{k}} \longleftarrow$ \textbf{\emph{Local Dataset Pruning}}
      \STATE \quad Incorporate prompts $p_{k}$ to the embedding of the inputs 
      \STATE \quad Forward propagation on model $W_{h,k}$ to get the smashed $S_{k}$. 
      \STATE \quad Send the smashed data $S_{k}$ to the server.
      \STATE\;

      \STATE \textbf{\emph{Client Backward Update}}
      \STATE \quad Get the smashed output ${S}_{k}^{'}$ from the server
      \STATE \quad Forward propagation on model $W_{t,k}$ 
      \STATE \quad Update $W_{t,k} = W_{t,k}-\eta \nabla \ell (W_{t,k},{S}_{k}^{'})$
      \STATE \quad Calculate the gradients $\hat{{S}_{k}}$ and send to the server.
      \STATE\;

      \STATE \textbf{\emph{Client Update}}
      \STATE \quad Wait for the gradients $\hat{S}_{k}$ from the server
      \STATE \quad Update $p_{k} = p_{k}-\eta \nabla \ell (p_{k},\hat{S}_{k})$
      \STATE \quad Send $W_{k}$ and $p_{k}$ to the server.
  \end{algorithmic}
\end{algorithm}

\begin{table*}[h]
\centering
\caption{The analysis of computational burden, communication cost, and latency of three methods in one global round }
\resizebox{\linewidth}{!}{
\begin{tabular}{@{}lccc@{}}
\toprule
\textbf{Methods} & \textbf{Computational Burden} & \textbf{Communication Cost } & \textbf{Latency}                                                                                                                                                                                          \\ \midrule
\multicolumn{1}{c|}{FL}           & \multicolumn{1}{c|}{$|D||W|$}                  & \multicolumn{1}{c|}{$2|W|K$}                  & $\frac{2|W|K}{R}+\frac{|D||W|U}{P_{C}}$                                                                                                                                                                                     \\
\multicolumn{1}{c|}{SFL}          & \multicolumn{1}{c|}{$(1-\tau)|D||W|$}            & \multicolumn{1}{c|}{$(4q|D|+2(1-\alpha-\tau)|W|)K$}    & $\frac{(4q|D|+2(1-\alpha-\tau)|W|)K}{R}+\frac{(1-\tau)|D||W|U}{P_{C}}+\frac{\tau|D||W|KU}{P_{S}}$                                                                                                                                 \\
\multicolumn{1}{l|}{SFPrompt}     & \multicolumn{1}{c|}{$(1-\tau)\gamma|D||W|$}            & \multicolumn{1}{c|}{$(4q\gamma|D|+2(1-\alpha-\tau)|W|)K$}     & $\frac{(2q\gamma|D|+2(1-\alpha-\tau)|W|) K}{R}+\frac{\alpha\beta\gamma|D||W|}{P_{C}}+
\max \left(\frac{(1-\tau)(1-\beta)\gamma|D||W|U}{P_{C}}, \frac{\tau\gamma|D||W| K}{P_{S}}+\frac{(1-\alpha-\tau)(1-\beta)\gamma|D||W|}{P_{C}}+\frac{2q\gamma|D|}{R}\right)$ \\ \bottomrule
\end{tabular}}
\label{analysis}
\end{table*}

\subsection{Phase 2: Split Training}
As training progresses, clients completing phase 1 introduce a set of $p$ continuous, learnable parameters, known as prompts, into the input space. The input space contains the embeddings of significant training samples following data pruning. Subsequently, the clients perform forward propagation using the model $W_{h}$ to generate `smashed' data, an intermediate output at the cut layer. The client then sends this smashed data to the server. On the sever-side, model $W_{b}$ performs forward propagation and transmits smashed data, the output of $W_{b}$, back to the client. Upon receiving the data, the client conducts both forward and backward propagation to update the tail model $W_{t}$ and compute the gradient.  The computed gradient is transmitted back to the server. After the server completes backward propagation, it transmits the gradient to the client, enabling the client to update the prompt $p$ based on the received gradient.
 
 Throughout the entire process, the prompt $p$ and tail $W_{t}$ are tuned, while the backbone $W_{b}$ and head $W_{h}$ remain frozen. This signifies that only a small subset of all parameters is being fine-tuned. Additionally, in this process, the raw data is always located on the local side, and there is no direct data sharing. The details of SFPrompt employed on the server side can be referred to Algorithm \ref{SFPrompt(server)}.

\subsection{Phase 3: Parameters Aggregation}
After completing the previous two phases, the parameters of the tail model $W_{t}$ and $p$ are updated. During this phase, clients transmit their updated local tail model and prompt parameters to the server, fostering a collaborative training process. The server subsequently conducts a global aggregation, then distributes the aggregated tail model and prompts the selected clients for the next round of training. This process is mathematically represented as follows:
\begin{equation}
(W_{t,r+1},p_{r+1})=\frac{1}{K}\sum_{k\in K} (W_{t,k,r},p_{k,r})
\end{equation}
where $r\in R$ is the number of global rounds, the final global model $W_{R}$ is derived by integrating the models $W_{h}$, $W_{b}$, $W_{t, R}$, and $p_{R}$ for inference. This integration results in the fine-tuned model, primed for downstream tasks. 

It's noteworthy to mention that SFPrompt achieves efficient tuning by only fine-tuning the tail model and prompt parameters. Importantly, this entire process occurs without sharing the raw data, thereby preserving data privacy. 


\begin{algorithm}[h]\small 
  \renewcommand{\algorithmicrequire}{\textbf{Input:}}
  \renewcommand{\algorithmicensure}{\textbf{Output:}}
  \caption{SFPrompt (Server)}
  \label{SFPrompt(server)}
  \begin{algorithmic}
      \REQUIRE ${K}$ denotes a set of selected clients at a given global round $r$, where $r \in R$, and $R$ is the number of global rounds. The smashed data of client $k$ at round $r$ is represented by $S_{k,r}$, and the prompt $p$ of client $k$ is represented as $p_{k}$. $N$ and $n_{k}$ stand for the total sample size and the sample size at a specific client $k$, respectively. $Y$ is the label set. Additionally, the number of local epochs is symbolized as $U$.
      \ENSURE $W_{R}$
      \STATE \;
      \STATE Split $W$ into three parts $W=[W_{h},W_{b},W_{t}]$.
      \STATE Keep the model $W_{b}$ frozen.
      \STATE Send $W_{h}$ and $W_{t}$ to the selected clients $K$
      \STATE \;
      \FOR{global round $r\in R$}   
      \FOR{each client $k\in K$ in parallel}
      \STATE $S_{k,r}\leftarrow$ \textbf{\emph{Client Forward Update}}
      \STATE Forward propagation with $S_{k,r}$ on $W_{b,r}$ and get $S_{k,r}^{'}$
      \STATE Send $S_{k,r}^{'}$ to the client
      \STATE Get the gradients $\hat{S}_{k,r}\leftarrow$ \textbf{\emph{Client Backward Update}}
      \STATE Calculate $\hat{S}_{k,r} = \nabla l_{k}(W_{b,r},\hat{S}_{k,r})$
      \STATE Send gradients $\hat{S}_{k,r}$ back client $k$
      \STATE $W_{t,k,r},p_{t,k,r}\longleftarrow \textbf{\emph{Client Update}}$
      \ENDFOR

      \STATE $(W_{t,r+1},p_{r+1})\longleftarrow \sum_{k=1}^{K} \frac{n_{k}}{N} (W_{t,k,r},p_{t,k,r})$
      \STATE Send the updated $W_{t,r+1}$ and $p_{r+1}$ to all $K$ clients
      \ENDFOR
      \STATE \;
      
      \RETURN Fine-tuned model $W_{R}=[W_{h,R},W_{b,R},W_{t,R}]$
  \end{algorithmic}
\end{algorithm}

\begin{figure*}
\centering
\includegraphics[width=\linewidth]{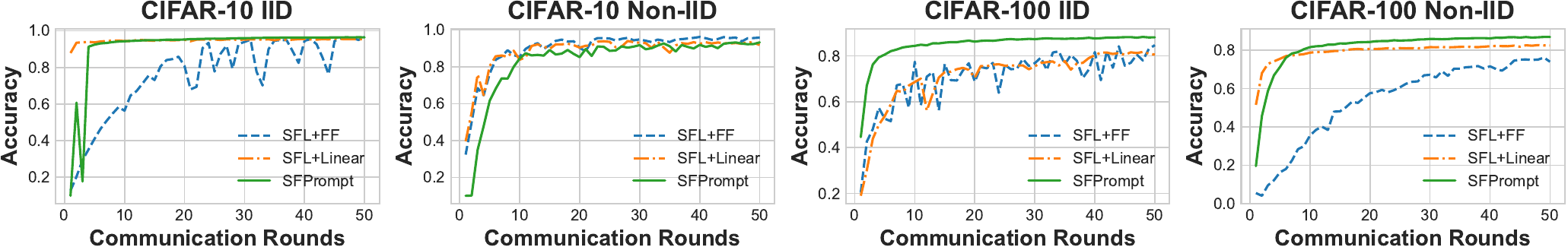}
\caption{Comparison of the accuracy among three methods. 
}
\label{main_results}
\end{figure*}

\subsection{Analysis of SFPrompt}
We delve into an in-depth analysis of SFPrompt. For easier understanding, we introduce the network split fraction, denoted as $\alpha$ and $\tau$. Then, we define that $|W_{h}|=\alpha|W|$, $|W_{b}|=\tau|W|$ and $|W_{t}|=(1-\alpha-\tau)|W|$, where $\left| W \right|$ signifies the total number of parameters in model $W$. We set the size of prompt parameters to $p$, the dataset pruning fraction to $\gamma$, and assume that the cut layer's size is $q$.
Further, we assume the computational power of the client and server as $P_{C}$ and $P_{S}$ respectively, with the condition that $P_{C} \ll P_{S}$. The time needed to update model $W$ on the dataset $D$ is expressed as $\frac{\left|D\right|\left|W\right|}{P}$, where forward propagation requires time $\beta\frac{\left|D\right|\left|W\right|}{P}$, and backward propagation demands time $(1-\beta)\frac{\left|D\right|\left|W\right|}{P}$. We have standardized uplink and downlink rates as $R$, which simplifies to $\frac{R}{K}$ when $K$ clients collaboratively work in order to streamline our analysis. We now proceed to analyze SFPrompt as following several perspectives. 

\noindent\textbf{Computational Cost.}
At the start of each global round, SFPrompt selects a set of $K$ clients to transmit the model $W_{C} = [W_{h}, W_{t}]$, causing a latency of $\frac{\left|W_{C}\right| K}{R}=\frac{(1-\tau) \left|W\right| K}{R}$. The selected clients then perform the forward propagation step, which requires a computation latency of $\frac{\gamma\beta(\left|W_{h}\right|+p)\left|D\right|}{P_{C}}=\frac{\gamma\beta(\alpha\left|W\right|+p)\left|D\right|}{P_{C}}$, and send the output of the cut layer to the server at the cost of $\frac{q\gamma\left|D\right|K}{R}$. Next, the server performs forward propagation on the received data, which equates to a computation of $\frac{\beta\tau\gamma\left|D\right|\left|W\right|}{P_{S}}$. Then send the output of the body to clients at the cost of $\frac{q\gamma\left|D\right|K}{R}$. The clients perform the forward and backward propagation on $W_{t}$, which requires a computation latency of $\frac{\gamma((1-\alpha-\tau)\left|W\right|+p)\left|D\right|}{P_{C}}$ and send the gradients back server costing $\frac{q\gamma\left|D\right|K}{R}$. Then the server performs the backward propagation costing $\frac{(1-\beta)\tau\gamma\left|D\right|\left|W\right|K}{P_{S}}$. 
Subsequently, the clients update their local model $W_{C}$ using the backward signal from the server, leading to additional computation costs of $\frac{\alpha(1-\beta)\gamma|D|(\left|W\right|+p)}{P_{C}}$. Finally, each client uploads the prompt $p$ and $W_{t}$ to the server for global aggregation, causing a latency of $\frac{(W_{t}+p)K}{R}$.


We compare SFPrompt with FL and SFL to analyze the efficiency and effectiveness, considering factors such as per-client computational burden, total communication cost, and overall latency, as detailed in Table \ref{analysis}. Primarily, both SFL and SFPrompt exhibit a reduced computational load due to their model-splitting strategy. Secondly, SFPrompt consistently outperforms SFL in terms of communication cost, as it capitalizes on local training to minimize frequent transmission. Lastly, our analysis shows that SFPrompt holds a significant advantage over FL when the model scale $W>\frac{2q\gamma}{\alpha+\tau}|D|$, making it an ideal choice for fine-tuning large models.  

\noindent \textbf{Privacy.}
SFPrompt maintains a privacy level consistent with other SFL schemes \cite{thapa2022splitfed} by keeping raw data localized. Nevertheless, akin to previously employed distributed learning methods, our algorithm might be susceptible to privacy concerns through model inversion attacks against the server, given that the server retains the parameters of the entire network. It's important to note that our work is orthogonal to existing privacy-preserving methods in SFL, such as protecting the intermediate activations \cite{vepakomma2020nopeek,titcombe2021practical}, and the incorporation of these methods can further fortify model privacy.

\section{Experiments}
\subsection{Experimental Setup\label{setup}}
\noindent \textbf{Pre-trained model and Downstream tasks.} 
In our experiments, we focus on vision fine-tuning tasks, employing ViT \cite{dosovitskiy2020image}, which is pre-trained on ImageNet-21k \cite{deng2009imagenet}. Specifically, we evaluate the performance of these models on four image classification datasets, namely, CIFAR-10 and CIFAR-100 \cite{krizhevsky2009learning}, SVHN \cite{netzer2011reading}, and Flower-102 \cite{nilsback2008automated}. 

\noindent \textbf{Baselines.} In our evaluation, we conduct a comprehensive comparison of SFPrompt with other commonly used fine-tuning methods to demonstrate its efficiency.
\begin{itemize}
  

    \item FL: This includes all traditional FL methods, such as FedSGD \cite{mcmahan2017communication}, which directly exchange the model to conduct the fine-tuning tasks.
    \item SFL \cite{thapa2022splitfed}: SFL splits the model and combines it with FL to train the model parallelly. 
    \item SFL+FF: This method combines SFL and full fine tuning (FF) that tunes all the model parameters.
    \item SFL+Linear: The method integrates SFL and Linear, which only fine-tune the linear layer while keeping the rest of the parameters frozen.
\end{itemize}

\noindent \textbf{Distributed Scenario Setting.} 
We construct a distributed learning scenario characterized by a central server endowed with substantial computational resources, coupled with 50 interconnected clients, each of which has limited computational capacity. At each round of the training process, only 5 clients are randomly selected to participate in the training and perform 10 local epochs in one global round. Data distribution across different clients falls into two types: IID (Independent and Identically Distributed) and Non-IID. The Non-IID data is split using a Dirichlet distribution \cite{hsu2019measuring} parameterized by $\alpha=0.1$.

\subsection{Evaluation of SFPrompt}
\begin{table}[]
\centering
\caption{The communication cost/ per round and computational burden/ per client of SFPrompt and other baselines}
\resizebox{\linewidth}{!}{
\begin{tabular}{@{}cccc@{}}
\toprule
Model                                                   & Methods  & Communication Cost & Computational Burden \\ \midrule
\multicolumn{1}{c|}{\multirow{3}{*}{\begin{tabular}[c]{@{}c@{}}ViT-Base\\ (391MB)\end{tabular}}}                 & FL       & 3910MB (1$\times$)                 & 16862.93GFLOPs ($1\times$)           \\ \cmidrule(l){2-4} 
\multicolumn{1}{c|}{}                                   & SFL      & 30380.86MB (7.77$\times$)           & 131.5GFLOPs (0.0078$\times$)                 \\ \cmidrule(l){2-4} 
\multicolumn{1}{c|}{}                                   & \textbf{SFPrompt} & \textbf{1825.19MB(0.47$\times$)}            & \textbf{78.9GFLOPs (0.0046$\times$)}         \\ \midrule
\multicolumn{1}{c|}{\multirow{3}{*}{\begin{tabular}[c]{@{}c@{}}ViT-Large\\ (1243MB)\end{tabular}}} & FL       & 12430MB (1$\times$)                    & 59685.79GFLOPs (1$\times$)                \\ \cmidrule(l){2-4} 
\multicolumn{1}{c|}{}                                   & SFL      & 40507.81MB (3.26$\times$)                         & 175.34GFLOPs (0.0029$\times$)                \\ \cmidrule(l){2-4} 
\multicolumn{1}{c|}{}                                   & \textbf{SFPrompt} & \textbf{2433.59MB (0.19$\times$)}                        & \textbf{105.2GFLOPs (0.0017$\times$)}                \\ \bottomrule
\end{tabular}}
\label{other metric}
\end{table}

\noindent \textbf{Accuracy.} We present the evaluation of SFPrompt's performance in various fine-tuning tasks across the chosen datasets. The comparison results among FF and linear are demonstrated in Fig \ref{main_results}. On the CIFAR-10 dataset, SFPrompt achieves performance comparable to baseline methods. The advancement of SFPrompt becomes more pronounced on the more complex CIFAR-100 dataset, where SFPrompt excelled entirely over the other two methods. 
Particularly in non-iid settings, the advantage of SFPrompt becomes even greater.
In this context, SFPrompt achieved a substantial 10.61\% improvement over FF and a 5.01\% improvement over Linear. 
We further extended our verification to the SVHN and Flower-102 datasets, where SFPrompt continued to demonstrate good performance. The detailed terminal results, showcasing the effectiveness of SFPrompt across different scenarios, are presented in Table \ref{Various dataset}.

\begin{table*}[]
\centering
\caption{The comparison of SFPrompt with baselines on various datasets}
\resizebox{\linewidth}{!}{
\begin{tabular}{@{}cccccccccc@{}}
\toprule
\multirow{2}{*}{Method}              & \multicolumn{2}{c}{CIFAR-10}                                                   & \multicolumn{2}{c}{CIFAR-100}                                                  & \multicolumn{2}{c}{SVHN}                                                      & \multicolumn{2}{c}{Flower-102}                                                & \multirow{2}{*}{\begin{tabular}[c]{@{}c@{}}Tuned Params \\ / Total Params\end{tabular}} \\ \cmidrule(lr){2-9}
                                     & \multicolumn{1}{c|}{IID}              & \multicolumn{1}{c|}{Non-IID}          & \multicolumn{1}{c|}{IID}              & \multicolumn{1}{c|}{Non-IID}          & \multicolumn{1}{c|}{IID}              & \multicolumn{1}{c|}{Non-IID}          & \multicolumn{1}{c|}{IID}              & \multicolumn{1}{c|}{Non-IID}          &                                                                                         \\ \midrule
\multicolumn{1}{c|}{SFL+FF} & \multicolumn{1}{c|}{\textbf{96.64\%}} & \multicolumn{1}{c|}{\textbf{95.63\%}} & \multicolumn{1}{c|}{84.71\%}          & \multicolumn{1}{c|}{76.16\%}          & \multicolumn{1}{c|}{81.38\%}          & \multicolumn{1}{c|}{71.83\%}          & \multicolumn{1}{c|}{94.61\%}          & \multicolumn{1}{c|}{90.55\%}          & 100\%                                                                                   \\ \midrule
\multicolumn{1}{c|}{SFL+Linear}          & \multicolumn{1}{c|}{95.62\%}          & \multicolumn{1}{c|}{94.03\%}          & \multicolumn{1}{c|}{82.67\%}          & \multicolumn{1}{c|}{81.76\%\%}        & \multicolumn{1}{c|}{79.19\%}          & \multicolumn{1}{c|}{71.73\%}          & \multicolumn{1}{c|}{93.56\%\%}        & \multicolumn{1}{c|}{90.87\%\%}        & \textbf{0.09\%}                                                                         \\ \midrule
\multicolumn{1}{c|}{SFPrompt}        & \multicolumn{1}{c|}{96.48\%}          & \multicolumn{1}{c|}{95.29\%}          & \multicolumn{1}{c|}{\textbf{88.23\%}} & \multicolumn{1}{c|}{\textbf{86.77\%}} & \multicolumn{1}{c|}{\textbf{82.29\%}} & \multicolumn{1}{c|}{\textbf{71.98\%}} & \multicolumn{1}{c|}{\textbf{94.78\%}} & \multicolumn{1}{c|}{\textbf{92.63\%}} & 0.18\%                                                                                  \\ \bottomrule
\end{tabular}}
\label{Various dataset}
\end{table*}

\noindent \textbf{Communication Cost.} We evaluate the communication cost of SFPrompt, comparing it with FL and SFL. The results, presented in Table \ref{other metric}, clearly illustrate that SFPrompt incurs lower communication costs compared to the other methods. This is due in part to SFPrompt performing local-loss updates, thereby eliminating the need for frequent data transmission and also pruning redundant examples from the local dataset to further reduce communication costs. SFPrompt's communication cost is 47\% that of FL and 6\% that of SFL. Intriguingly, the gap between these methods grows as the model size increases, making the efficiency improvements more pronounced. Using the ViT-large with more parameters as the base model, the communication cost for SFPrompt is reduced to 19\% that of FL. This disparity stems from FL's communication cost being tied to the model size, while SFPrompt's cost is associated with the number of interactions and the size of the transmitted data, having little correlation with the model size. 


\noindent \textbf{Computational Burden.} We assess the computational cost in terms of FLOPs, a widely accepted measure of model complexity. The results reveal that both SFPrompt and SFL simplify local model computation. This simplification is achieved by splitting the model to position the simpler model on clients, which is with limited computational resources. The computational complexity of SFPrompt is a mere 0.46\% that of FL, and this advantage magnifies as the model size increases. With the model size increasing, SFPrompt manages to maintain a simple model structure on the client side. Although the client model of SFPrompt and SFL is identical, SFPrompt utilizes fewer samples, thereby reducing FLOPs.

\begin{figure}
    \centering
    \includegraphics[width=\linewidth]{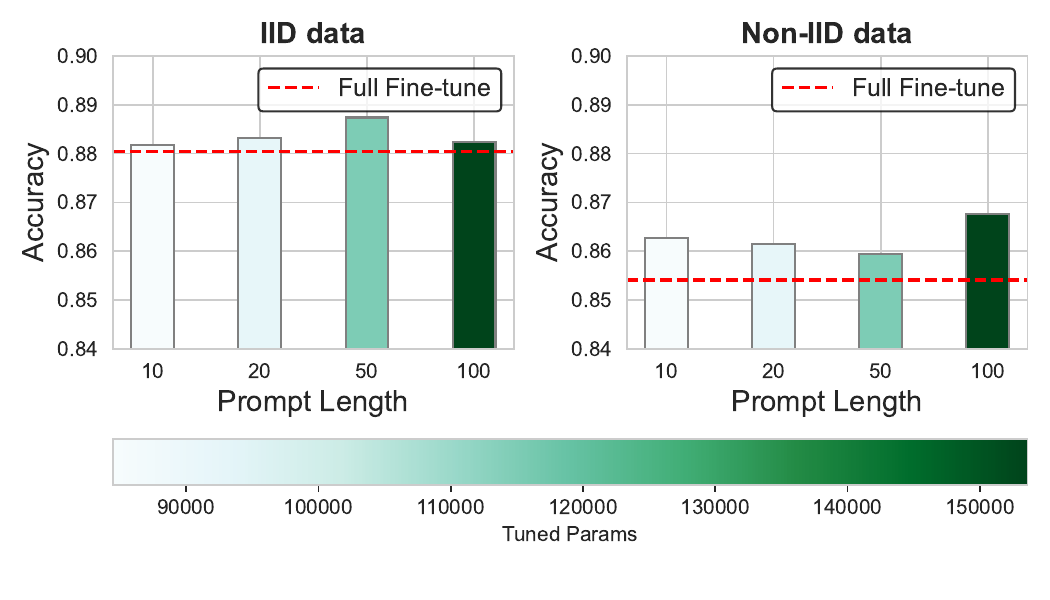}
    \caption{The accuracy of SFPrompt with various prompt lengths and tuned parameters on CIFAR-100}
    \label{prompt_length }
\end{figure}

\noindent \textbf{Prompt Length.} We investigate the influence of the hyperparameter pertaining to the prompt length on performance. The prompt length specifically governs the number of parameters fine-tuned within SFPrompt. For various prompt lengths, the corresponding adjusted parameters and model performance are depicted in Figure \ref{prompt_length }. It becomes evident that the optimal prompt length is not uniform but varies across different tasks. Upon comparing with other methods in Table \ref{Various dataset}, we find that SFPrompt necessitates fewer tuned parameters yet delivers competitive performance.

\subsection{Ablation Study}
\begin{figure}
    \centering
    \includegraphics[width=0.85\linewidth]{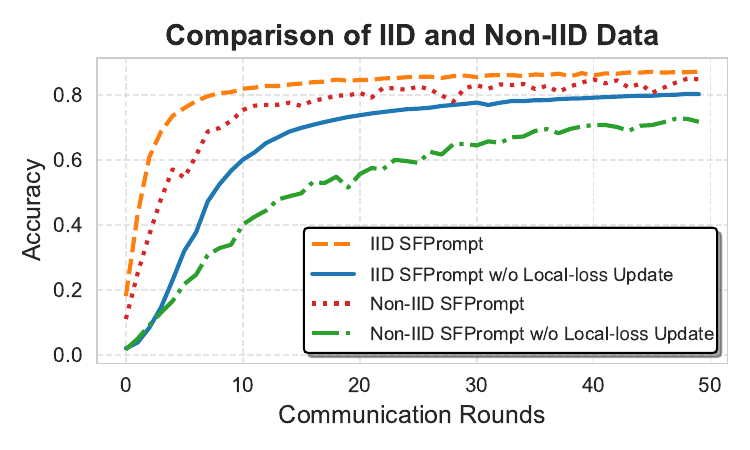}
    \caption{The performance of SFPrompt and SFPrompt w/o local-loss update on CIFAR-100}
    \label{local update}
\end{figure}

\textbf{Local-loss Update.} To further understand the effectiveness and robustness of SFPrompt, we conduct an ablation study, exploring the influence of different components and configurations on the model's performance. 
Firstly, we compare SFPrompt with SFPrompt w/o local-loss update to explore the influence of local-loss update, as shown in Figure \ref{local update}. The results indicate that the local-loss update step is instrumental in contributing to SFPrompt's performance.

\begin{figure}
    \centering
    \includegraphics[width=\linewidth]{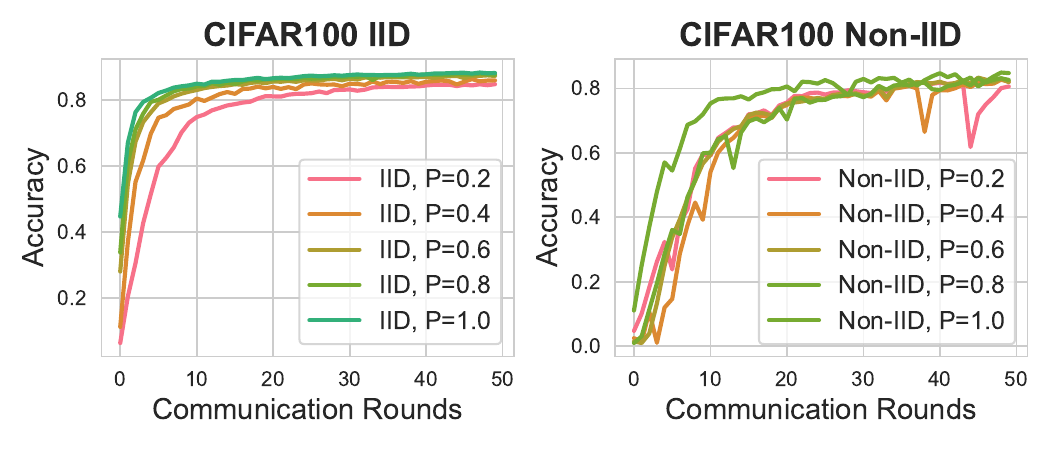}
    \caption{The accuracy of SFPrompt under different local dataset pruning fraction}
    \label{dataset_pruning}
\end{figure}

\noindent \textbf{Dataset Pruning.} We evaluate the effect of different local dataset pruning fractions on the performance of the model, as depicted in Figure \ref{dataset_pruning}. Interestingly, even with deep pruning of the local dataset under the IID condition, the impact on model performance was minimal. When only 20\% of the largest EL2N values in the local dataset were retained, the performance was only reduced by 3.39\% compared to the full dataset. Under non-IID conditions, even after pruning 80\% of the data, the performance decline is quite limited, amounting to only 4.32\%. This occurs because, although SFPrompt doesn't utilize the complete dataset during global training, it leverages the complete data in subsequent local-loss updates. Additionally, in a distributed scenario, data from other clients can still provide rich information and compensate for losses to support the training process.

\section{Conclusions}
In this paper, SFPrompt is introduced as a privacy-preserving and efficient distributed fine-tuning framework.
SFPrompt splits the large pre-trained model into server-side and client-side to lower the computational burden on the client and further introduce prompt parameters to enable efficient fine-tuning.
SFPrompt prunes the redundant local data and updates the model based on the local loss to further reduce the additional communication cost.
Our extensive experiments reveal that SFPrompt achieves state-of-the-art performance, outperforming existing baselines.

\bibliographystyle{named}
\bibliography{ijcai24}

\begin{thebibliography}{}

\bibitem[\protect\citeauthoryear{Brown \bgroup \em et al.\egroup }{2020}]{brown2020language}
Tom Brown, Benjamin Mann, Nick Ryder, Melanie Subbiah, Jared~D Kaplan, Prafulla Dhariwal, Arvind Neelakantan, Pranav Shyam, Girish Sastry, Amanda Askell, et~al.
\newblock Language models are few-shot learners.
\newblock {\em Proc. NeurIPS}, 33:1877--1901, 2020.

\bibitem[\protect\citeauthoryear{Chen \bgroup \em et al.\egroup }{2022}]{chen2022fedtune}
Jinyu Chen, Wenchao Xu, Song Guo, Junxiao Wang, Jie Zhang, and Haozhao Wang.
\newblock Fedtune: A deep dive into efficient federated fine-tuning with pre-trained transformers.
\newblock {\em arXiv preprint arXiv:2211.08025}, 2022.

\bibitem[\protect\citeauthoryear{Dale}{2021}]{dale2021gpt}
Robert Dale.
\newblock Gpt-3: What’s it good for?
\newblock {\em Natural Language Engineering}, 27(1):113--118, 2021.

\bibitem[\protect\citeauthoryear{Deng \bgroup \em et al.\egroup }{2009}]{deng2009imagenet}
Jia Deng, Wei Dong, Richard Socher, Li-Jia Li, Kai Li, and Li~Fei-Fei.
\newblock Imagenet: A large-scale hierarchical image database.
\newblock In {\em Proc. CVPR}, pages 248--255, 2009.

\bibitem[\protect\citeauthoryear{Desislavov \bgroup \em et al.\egroup }{2021}]{desislavov2021compute}
Radosvet Desislavov, Fernando Mart{\'\i}nez-Plumed, and Jos{\'e} Hern{\'a}ndez-Orallo.
\newblock Compute and energy consumption trends in deep learning inference.
\newblock {\em arXiv preprint arXiv:2109.05472}, 2021.

\bibitem[\protect\citeauthoryear{Dosovitskiy \bgroup \em et al.\egroup }{2020}]{dosovitskiy2020image}
Alexey Dosovitskiy, Lucas Beyer, Alexander Kolesnikov, Dirk Weissenborn, Xiaohua Zhai, Thomas Unterthiner, Mostafa Dehghani, Matthias Minderer, Georg Heigold, Sylvain Gelly, et~al.
\newblock An image is worth 16x16 words: Transformers for image recognition at scale.
\newblock {\em arXiv preprint arXiv:2010.11929}, 2020.

\bibitem[\protect\citeauthoryear{Guo \bgroup \em et al.\egroup }{2023}]{guo2023promptfl}
Tao Guo, Song Guo, Junxiao Wang, Xueyang Tang, and Wenchao Xu.
\newblock Promptfl: Let federated participants cooperatively learn prompts instead of models-federated learning in age of foundation model.
\newblock {\em IEEE TMC}, 2023.

\bibitem[\protect\citeauthoryear{Gupta and Raskar}{2018}]{gupta2018distributed}
Otkrist Gupta and Ramesh Raskar.
\newblock Distributed learning of deep neural network over multiple agents.
\newblock {\em Journal of Network and Computer Applications}, 116:1--8, 2018.

\bibitem[\protect\citeauthoryear{Houlsby \bgroup \em et al.\egroup }{2019}]{houlsby2019parameter}
Neil Houlsby, Andrei Giurgiu, Stanislaw Jastrzebski, Bruna Morrone, Quentin De~Laroussilhe, Andrea Gesmundo, Mona Attariyan, and Sylvain Gelly.
\newblock Parameter-efficient transfer learning for nlp.
\newblock In {\em Proc. ICML}, pages 2790--2799. PMLR, 2019.

\bibitem[\protect\citeauthoryear{Hsu \bgroup \em et al.\egroup }{2019}]{hsu2019measuring}
Tzu-Ming~Harry Hsu, Hang Qi, and Matthew Brown.
\newblock Measuring the effects of non-identical data distribution for federated visual classification.
\newblock {\em arXiv preprint arXiv:1909.06335}, 2019.

\bibitem[\protect\citeauthoryear{Hu \bgroup \em et al.\egroup }{2021}]{hu2021lora}
Edward~J Hu, Yelong Shen, Phillip Wallis, Zeyuan Allen-Zhu, Yuanzhi Li, Shean Wang, Lu~Wang, and Weizhu Chen.
\newblock Lora: Low-rank adaptation of large language models.
\newblock {\em arXiv preprint arXiv:2106.09685}, 2021.

\bibitem[\protect\citeauthoryear{Jia \bgroup \em et al.\egroup }{2022}]{jia2022visual}
Menglin Jia, Luming Tang, Bor-Chun Chen, Claire Cardie, Serge Belongie, Bharath Hariharan, and Ser-Nam Lim.
\newblock Visual prompt tuning.
\newblock In {\em Proc. ECCV}, pages 709--727. Springer, 2022.

\bibitem[\protect\citeauthoryear{Kairouz \bgroup \em et al.\egroup }{2021}]{kairouz2021advances}
Peter Kairouz, H~Brendan McMahan, Brendan Avent, Aur{\'e}lien Bellet, Mehdi Bennis, Arjun~Nitin Bhagoji, Kallista Bonawitz, Zachary Charles, Graham Cormode, Rachel Cummings, et~al.
\newblock Advances and open problems in federated learning.
\newblock {\em Foundations and Trends{\textregistered} in Machine Learning}, 14(1--2):1--210, 2021.

\bibitem[\protect\citeauthoryear{Kone{\v{c}}n{\`y} \bgroup \em et al.\egroup }{2016}]{konevcny2016federated}
Jakub Kone{\v{c}}n{\`y}, H~Brendan McMahan, Felix~X Yu, Peter Richt{\'a}rik, Ananda~Theertha Suresh, and Dave Bacon.
\newblock Federated learning: Strategies for improving communication efficiency.
\newblock {\em arXiv preprint arXiv:1610.05492}, 2016.

\bibitem[\protect\citeauthoryear{Krizhevsky \bgroup \em et al.\egroup }{2009}]{krizhevsky2009learning}
Alex Krizhevsky, Geoffrey Hinton, et~al.
\newblock Learning multiple layers of features from tiny images.
\newblock 2009.

\bibitem[\protect\citeauthoryear{McMahan \bgroup \em et al.\egroup }{2017}]{mcmahan2017communication}
Brendan McMahan, Eider Moore, Daniel Ramage, Seth Hampson, and Blaise~Aguera y~Arcas.
\newblock Communication-efficient learning of deep networks from decentralized data.
\newblock In {\em Proc. AISTATS}, pages 1273--1282. PMLR, 2017.

\bibitem[\protect\citeauthoryear{Netzer \bgroup \em et al.\egroup }{2011}]{netzer2011reading}
Yuval Netzer, Tao Wang, Adam Coates, Alessandro Bissacco, Bo~Wu, and Andrew~Y Ng.
\newblock Reading digits in natural images with unsupervised feature learning.
\newblock 2011.

\bibitem[\protect\citeauthoryear{Nilsback and Zisserman}{2008}]{nilsback2008automated}
Maria-Elena Nilsback and Andrew Zisserman.
\newblock Automated flower classification over a large number of classes.
\newblock In {\em 2008 Sixth Indian conference on computer vision, graphics \& image processing}, pages 722--729, 2008.

\bibitem[\protect\citeauthoryear{Niu \bgroup \em et al.\egroup }{2022}]{niu2022federated}
Yue Niu, Saurav Prakash, Souvik Kundu, Sunwoo Lee, and Salman Avestimehr.
\newblock Federated learning of large models at the edge via principal sub-model training.
\newblock {\em arXiv preprint arXiv:2208.13141}, 2022.

\bibitem[\protect\citeauthoryear{Paul \bgroup \em et al.\egroup }{2021}]{paul2021deep}
Mansheej Paul, Surya Ganguli, and Gintare~Karolina Dziugaite.
\newblock Deep learning on a data diet: Finding important examples early in training.
\newblock {\em Proc. NeurIPS}, 34:20596--20607, 2021.

\bibitem[\protect\citeauthoryear{Thapa \bgroup \em et al.\egroup }{2022}]{thapa2022splitfed}
Chandra Thapa, Pathum Chamikara~Mahawaga Arachchige, Seyit Camtepe, and Lichao Sun.
\newblock Splitfed: When federated learning meets split learning.
\newblock In {\em Proc. AAAI}, pages 8485--8493, 2022.

\bibitem[\protect\citeauthoryear{Titcombe \bgroup \em et al.\egroup }{2021}]{titcombe2021practical}
Tom Titcombe, Adam~J Hall, Pavlos Papadopoulos, and Daniele Romanini.
\newblock Practical defences against model inversion attacks for split neural networks.
\newblock {\em arXiv preprint arXiv:2104.05743}, 2021.

\bibitem[\protect\citeauthoryear{Vepakomma \bgroup \em et al.\egroup }{2018}]{vepakomma2018split}
Praneeth Vepakomma, Otkrist Gupta, Tristan Swedish, and Ramesh Raskar.
\newblock Split learning for health: Distributed deep learning without sharing raw patient data.
\newblock {\em arXiv preprint arXiv:1812.00564}, 2018.

\bibitem[\protect\citeauthoryear{Vepakomma \bgroup \em et al.\egroup }{2020}]{vepakomma2020nopeek}
Praneeth Vepakomma, Abhishek Singh, Otkrist Gupta, and Ramesh Raskar.
\newblock Nopeek: Information leakage reduction to share activations in distributed deep learning.
\newblock In {\em Proc. ICDMW)}, pages 933--942, 2020.

\bibitem[\protect\citeauthoryear{Voigt and Von~dem Bussche}{2017}]{voigt2017eu}
Paul Voigt and Axel Von~dem Bussche.
\newblock The eu general data protection regulation (gdpr).
\newblock {\em A Practical Guide, 1st Ed., Cham: Springer International Publishing}, 10(3152676):10--5555, 2017.

\bibitem[\protect\citeauthoryear{Wang \bgroup \em et al.\egroup }{2023}]{wang2023large}
Xiao Wang, Guangyao Chen, Guangwu Qian, Pengcheng Gao, Xiao-Yong Wei, Yaowei Wang, Yonghong Tian, and Wen Gao.
\newblock Large-scale multi-modal pre-trained models: A comprehensive survey.
\newblock {\em Machine Intelligence Research}, pages 1--36, 2023.

\bibitem[\protect\citeauthoryear{Xiao \bgroup \em et al.\egroup }{2023}]{xiao2023offsite}
Guangxuan Xiao, Ji~Lin, and Song Han.
\newblock Offsite-tuning: Transfer learning without full model.
\newblock {\em arXiv preprint arXiv:2302.04870}, 2023.

\bibitem[\protect\citeauthoryear{Zhang \bgroup \em et al.\egroup }{2022}]{zhang2022federated}
Zhuo Zhang, Yuanhang Yang, Yong Dai, Lizhen Qu, and Zenglin Xu.
\newblock When federated learning meets pre-trained language models' parameter-efficient tuning methods.
\newblock {\em arXiv preprint arXiv:2212.10025}, 2022.

\bibitem[\protect\citeauthoryear{Zhao \bgroup \em et al.\egroup }{2023}]{zhao2023fedprompt}
Haodong Zhao, Wei Du, Fangqi Li, Peixuan Li, and Gongshen Liu.
\newblock Fedprompt: Communication-efficient and privacy-preserving prompt tuning in federated learning.
\newblock In {\em Proc. ICASSP 2023}, pages 1--5, 2023.

\end{thebibliography}

\end{document}


\maketitle

\clearpage
\section{Appendix}
\subsection{The Difference between Existing Methods}
Recent works have explored the application of federated learning (FL) to large-scale model fine-tuning, yet there are notable differences from the approach described in this work. The existing methods can broadly be categorized into two main streams. The first involves a direct combination of FL and parameter-efficient fine-tuning \cite{zhao2023fedprompt,guo2023promptfl,zhang2022federated,zhang2023towards,chen2022fedtune,ding2023parameter,li2023visual,chen2023prompt}, where each client is assumed to own a pre-trained model that uses local data for fine-tuning before aggregating the fine-tuned parameters. However, this approach conducts the entire fine-tuning process locally, overlooking the possibility that local resources may not be sufficient to support model fine-tuning. For instance, a single inference with GPT-3 requires 740 TFLOPs, a demand that is beyond the reach of typical consumer devices. The second approach focuses on combining parameter-efficient fine-tuning and the model emulator. This strategy considers lossy compression of the original large model through distillation, quantization, or pruning \cite{xiao2023offsite,niu2022federated,chen2022fedobd}. Fine-tuning is achieved by transmitting the compressed model and incorporating it with parameter-efficient methods. Yet, the compressed models resulting from this method often remain substantial in size, imposing higher demands on both communication costs and local computing resources. An example of this can be found in work \cite{xiao2023offsite}, where layer retention is determined by a specific stride. With models like GPT-3, the compressed version remains quite large, failing to fully address the aforementioned challenges.

SFPrompt introduces a solution for fine-tuning in distributed environments by integrating SFL and prompt learning (PT). SFL can significantly alleviate local computational burdens while ensuring data privacy. SFPrompt leverages PT for efficient fine-tuning. Since every round of SFL necessitates interaction with a server, SFPrompt introduces strategies like local-loss update and dataset pruning to further reduce communication costs. As the model size grows, the advantages of SFPrompt in reducing communication expenses and local computational demands continue to amplify.

\subsection{Implementation Details}
We run all experiments on a 24GB NVIDIA RTX3090 GPU. We used Pytorch and Timm in our implementations.

\begin{figure}
    \centering
    \includegraphics[width=\linewidth]{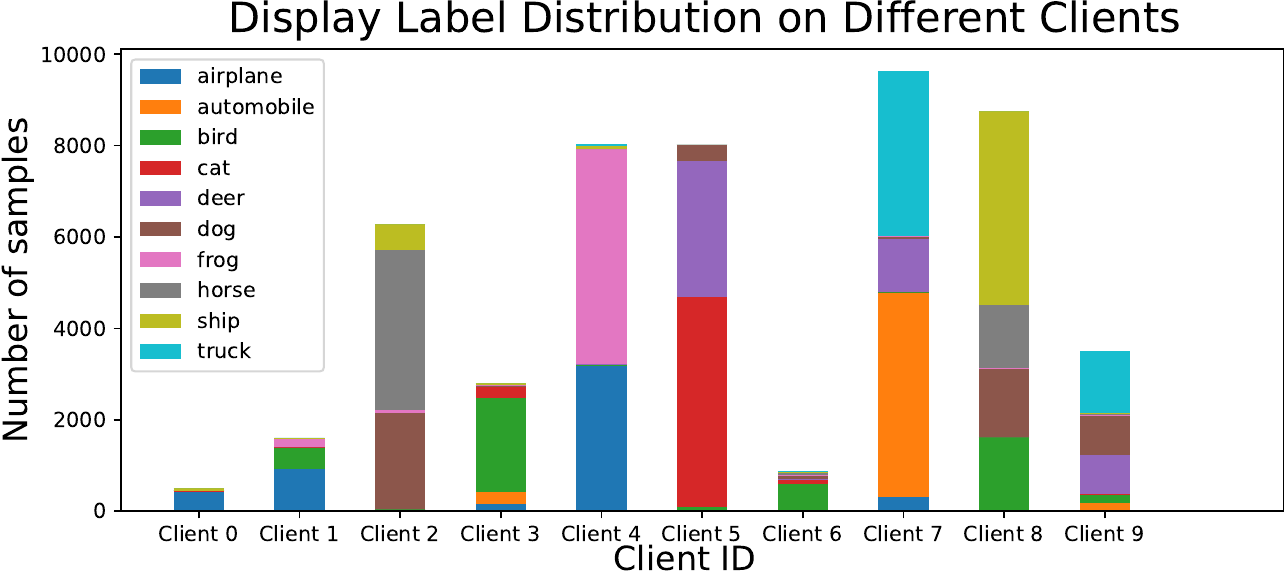}
    \caption{The Non-IID distribution ($\alpha=0.1$) of data across different clients}
    \label{non-iid}
\end{figure}

We use CIFAR-10, CIFAR-100\cite{krizhevsky2009learning}, SVHN \cite{netzer2011reading}, and Flower-102\cite{nilsback2008automated}. The division of the training set and test set follows the default setting. To construct the distributed setting, we further divide the training set according to the number of clients and the data distribution. Figure \ref{non-iid} is the data distribution of different clients of CIFAR10 under the Non-IID setting. The test set is always on the server side, testing the performance of the complete $W=[W_{S}, W_{C}]$. 

We use the ViT as the foundation model, which is pre-trained on Imagenet-21K \cite{deng2009imagenet}. All networks are trained with Stochastic Gradient Descent (SGD) optimizer, the global training learning rate is 0.1, and the local update learning rate is set to $10^{-4}$. For all datasets, the batch size is set to 128. 

\subsection{Additional Experiments}
\begin{figure}
    \centering
    \includegraphics[width=\linewidth]{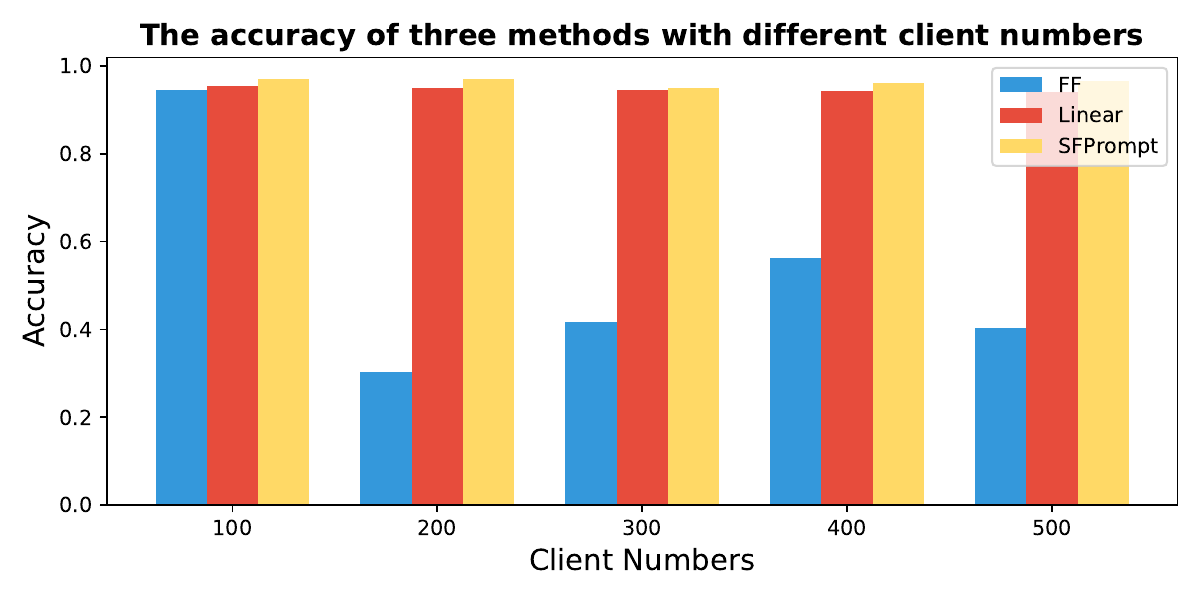}
    \caption{The performance of three fine-tuning method with different number of clients}
    \label{client number}
\end{figure}

\textbf{Number of clients.}We conduct experiments involving various numbers of clients, as depicted in Figure \ref{client number}. The training set, comprising a total of 50,000 images from CIFAR-10, is divided according to the number of clients. With an increase in the number of clients, the quantity of images assignable to each client diminishes. Figure \ref{client number} illustrates that SFPrompt demonstrates remarkable robustness as the client count grows, consistently outperforming the other two methods. FF performs poorly in such an experimental setting, as it is difficult for FF to achieve a better result when there is little local data on a single client, possibly leading to overfitting. Linear has always performed well, but still not as good as SFPrompt.

\begin{figure}
    \raggedright
    \includegraphics[width=0.48\linewidth]{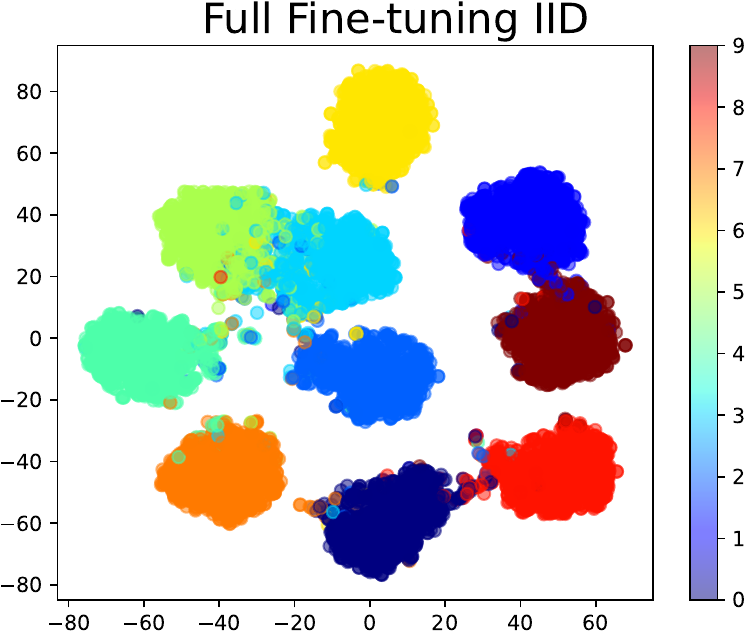}
    \includegraphics[width=0.48\linewidth]{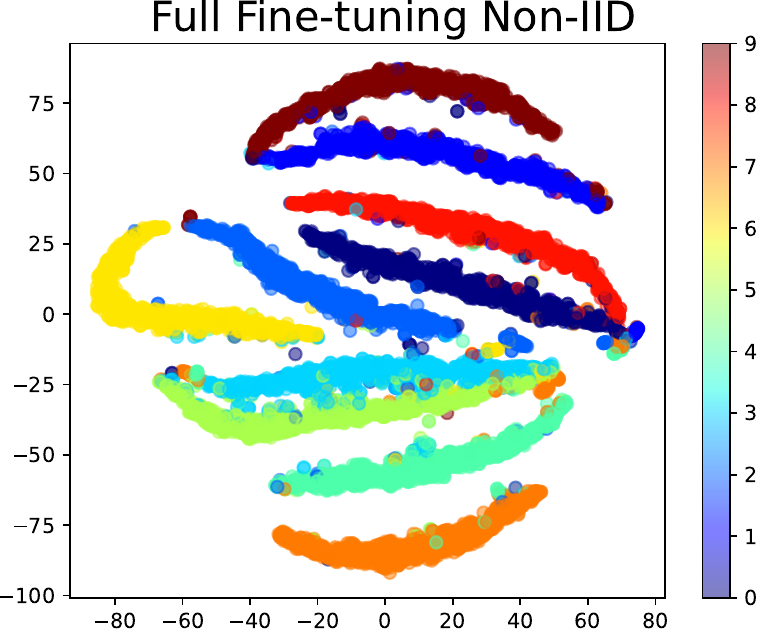}
    \caption{T-SNE visualization of Full Fine-tune on CIFAR-10}
    \label{ff}
\end{figure}

\begin{figure}
    \raggedright
    \includegraphics[width=0.48\linewidth]{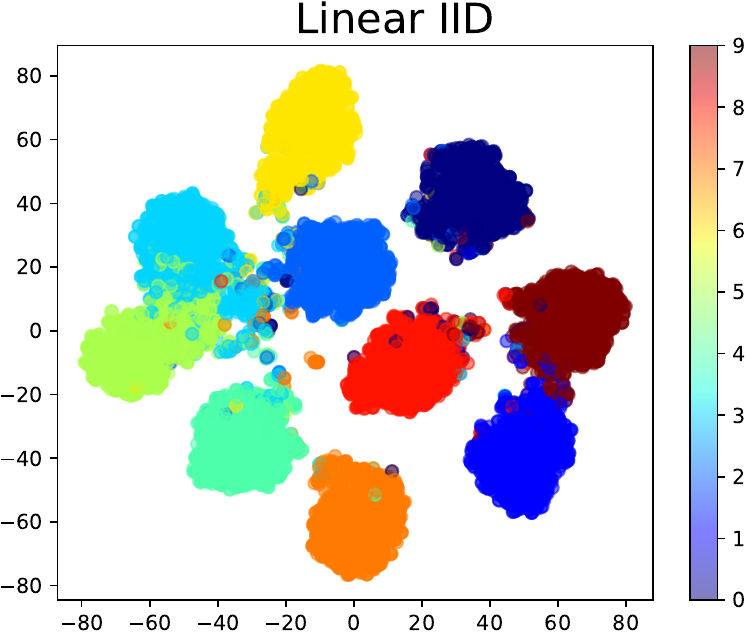}
    \includegraphics[width=0.48\linewidth]{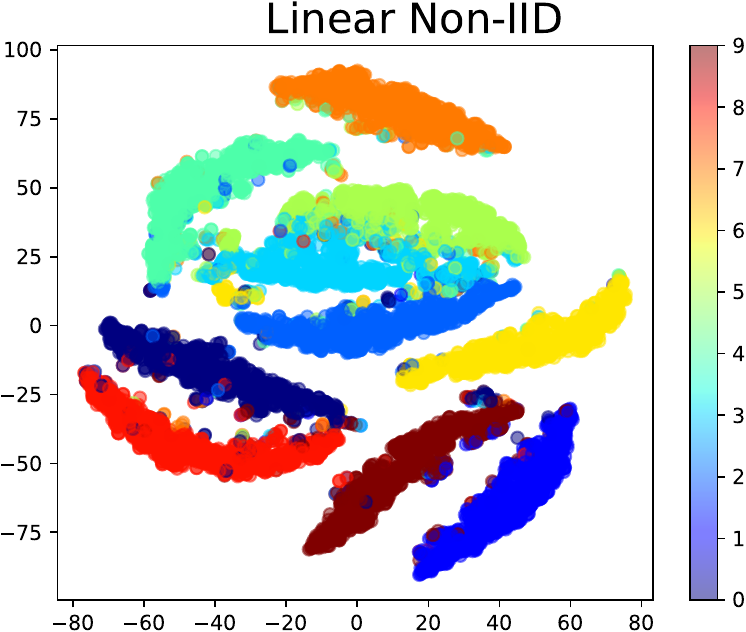}
    \caption{T-SNE visualization of Linear on CIFAR-10}
    \label{linear}
\end{figure}

\begin{figure}
    \raggedright
    \includegraphics[width=0.48\linewidth]{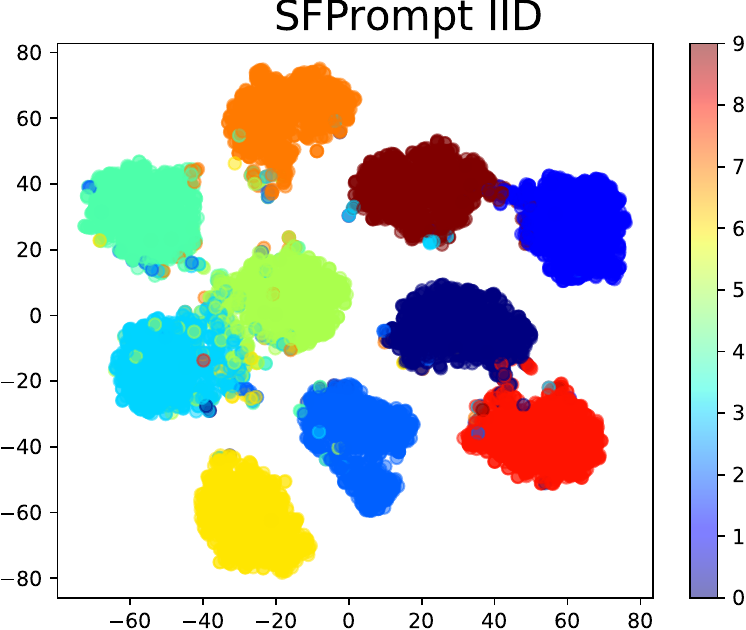}
    \includegraphics[width=0.48\linewidth]{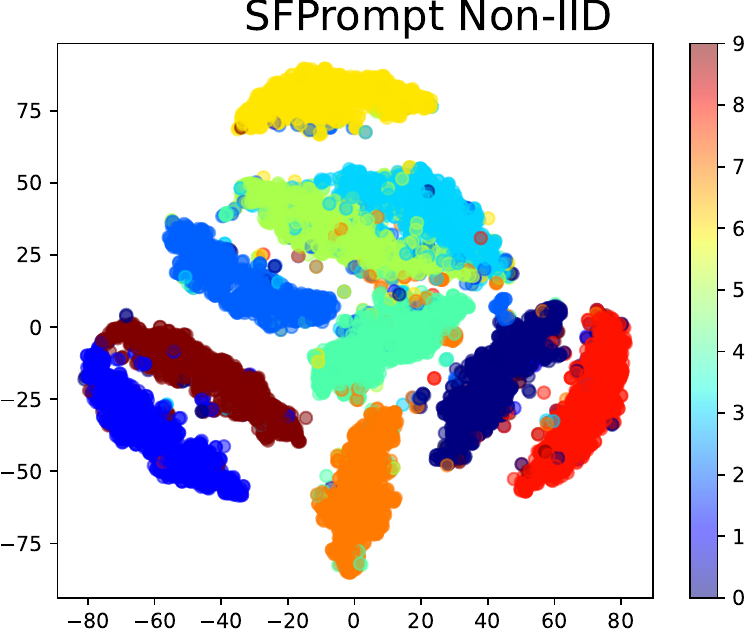}
    \caption{T-SNE visualization of SFPrompt on CIFAR-10}
    \label{SFPrompt}
\end{figure}

\noindent \textbf{Visualization.} We present the t-SNE visualization of SFPrompt in Figure \ref{SFPrompt} and juxtapose it with the full fine-tuning in Fig. \ref{ff}, linear fine-tuning in Fig. \ref{linear}, and the original model without fine-tuning shown in Fig. \ref{vit}. The figure shows t-SNE visualizations of the embedding after the tail model on CIFAR-10. The outcomes highlight SFPrompt's capability to generate linearly separable features without updating all backbone parameters, contrasting with full fine-tuning, thus exemplifying parameter-efficient fine-tuning.

\noindent \textbf{Runtime Estimate.} We offer mathematical estimates (see Table \ref{run-time}) using network speed data from \cite{guo2023promptfl} and train latency from \cite{jia2022visual}. Theoretically, SFPrompt is expected to save more time. However, in real-world scenarios, factors like network quality introduce complexity, warranting further exploration in future work.

\begin{table}[htbp]
\centering
\resizebox{1\linewidth}{!}{
\begin{tabular}{@{}c|c|cc@{}}
\toprule
Method                          & FL        & \multicolumn{1}{c|}{SFL} & SFPrompt  \\ \midrule
Run-time (54 Mbps ↓, 12 Mbps ↑) & 4.091 min & 4.648 min                           & \textbf{3.505 min} \\ \bottomrule
\end{tabular}}
\caption{Mathematical estimation of running time}
\label{run-time}
\end{table}

\noindent \textbf{Other dataset.}
To more convincingly demonstrate the superiority of SFPrompt, we conducted experiments on Caltech 101 \cite{li_andreeto_ranzato_perona_2022} and Food101 \cite{bossard14}, with the results presented in Table \ref{caltech and food}. It is evident that SFPrompt consistently exhibits better performance.

\begin{table}[htbp]
\centering
\resizebox{0.9\linewidth}{!}{
\begin{tabular}{@{}ccccc@{}}
\toprule
\multirow{2}{*}{Method}       & \multicolumn{2}{c}{Caltech 101}                             & \multicolumn{2}{c}{Food 101}           \\ \cmidrule(l){2-5} 
                              & \multicolumn{1}{c|}{IID}     & \multicolumn{1}{c|}{Non-IID} & \multicolumn{1}{c|}{IID}     & Non-IID \\ \midrule
\multicolumn{1}{c|}{SFL+Linear}   & \multicolumn{1}{c|}{92.62\%} & \multicolumn{1}{c|}{90.43\%} & \multicolumn{1}{c|}{79.85\%} & 75.53\% \\ \midrule
\multicolumn{1}{c|}{SFPrompt} & \multicolumn{1}{c|}{94.04\%} & \multicolumn{1}{c|}{92.18\%} & \multicolumn{1}{c|}{85.31\%} & 80.11\% \\ \bottomrule
\end{tabular}}
\caption{Additional experiments on Caltech 101 and Food 101}
\label{caltech and food}
\end{table}

\begin{figure}
    \raggedright
    \includegraphics[width=0.48\linewidth]{Pics/Pre-trained model.pdf}
    \caption{T-SNE visualization of ViT without fine-tuning on CIFAR-10}
    \label{vit}
\end{figure}


\bibliographystyle{named}
\bibliography{appendix.bib}
\vfill